\documentclass[review]{elsarticle}

\usepackage{lineno,hyperref}
\modulolinenumbers[5]
\usepackage{times}
\usepackage{epsfig}
\usepackage{graphicx}
\usepackage{amsmath}
\usepackage{amssymb}
\usepackage{multirow}
\usepackage{array}
\usepackage{bm}
\usepackage{tabularx}
\usepackage{amsmath}
\usepackage{amssymb}
\usepackage{latexsym}
\usepackage{algpseudocode}
\usepackage{tabularx}
\usepackage{booktabs}
\usepackage{url}

\usepackage{algorithmicx}
\usepackage{algorithm}
\usepackage{graphicx}
\usepackage{amssymb,subfigure,cases}
\usepackage{times}
\usepackage{color}
\usepackage{multirow}
\usepackage{algpseudocode}
\usepackage{amsmath}
\usepackage{algorithmicx}
\usepackage{algorithm}
\usepackage{graphicx}
\usepackage{tabularx}
\usepackage{amssymb,subfigure,cases}
\usepackage{times}
\usepackage{color}
\usepackage{multirow}
\usepackage{array}

\journal{Journal of \LaTeX\ Templates}

\begin{document}
\begin{frontmatter}
\title{Hierarchical Multimodal Transformer to Summarize Videos}

\author[1,3]{Bin Zhao}
\author[2]{Maoguo~Gong\corref{c}}\cortext[c]{Corresponding author}\ead{gong@ieee.org}

\author[3]{Xuelong Li}

\address[1]{Academy of Advanced Interdisciplinary Research, Xidian University, Xi’an 710071, China.}
\address[2]{School of Electronic Engineering, and the Key Laboratory of Intelligent Perception and Image Understanding, Xidian University, Xi’an 710071, China.}
\address[3]{School of Artificial Intelligence, Optics and Electronics (iOPEN), Northwestern Polytechnical University, Xi'an 710072, China.}

\begin{abstract}

Although video summarization has achieved tremendous success benefiting from Recurrent Neural Networks (RNN), RNN-based methods neglect the global dependencies and multi-hop relationships among video frames, which limits the performance. Transformer is an effective model to deal with this problem, and surpasses RNN-based methods in several sequence modeling tasks, such as machine translation, video captioning, \emph{etc}. Motivated by the great success of transformer and the natural structure of video (frame-shot-video), a hierarchical transformer is developed for video summarization, which can capture the dependencies among frame and shots, and summarize the video by exploiting the scene information formed by shots. Furthermore, we argue that both the audio and visual information are essential for the video summarization task. To integrate the two kinds of information, they are encoded in a two-stream scheme, and a multimodal fusion mechanism is developed based on the hierarchical transformer. In this paper, the proposed method is denoted as Hierarchical Multimodal Transformer (HMT). Practically, extensive experiments show that HMT achieves (F-measure: 0.441, Kendall's \(\tau\): 0.079, Spearman's \(\rho\): 0.080) and (F-measure: 0.601, Kendall's \(\tau\): 0.096, Spearman's \(\rho\): 0.107) on SumMe and TVsum, respectively. It surpasses most of the traditional, RNN-based and attention-based video summarization methods.
\end{abstract}

\begin{keyword}
transformer, multimodal fusion, video summarization, hierarchical structure
\end{keyword}
\end{frontmatter}

\section{Introduction}\label{section1}

Video has become the most popular data format in our daily life, including communication, interaction and entertainment. The wide applications lead to the explosion of video data. Statistics show that video data has occupied over 90\% of current Internet traffic \cite{estevam2021zero,tian2017unified}. It also takes the leading position in mobile Internet, and still keeps growing rapidly. Moreover, it is widely used in the tasks of artificial intelligence, such as automatic driving, intelligent monitoring, and interactive robot \cite{peng2021deep}. Under this circumstance, the automatic video analysis techniques are demanded urgently. Video summarization is one of the techniques rising for the explosion of video data \cite{li2017multiview,sahu2020summarizing}. It highlights the main video content and removes redundancy by selecting a subset of key-frames or key-shots, denoted as video summary. In this paper, we focus on the key-shot based video summary, which is a brief version of the original video. It can ease the way of viewers to browse the video content, and can also boost the efficiency of other video-based computer vision tasks, such as video captioning, action recognition, anomaly detection, \emph{etc.} \cite{DBLP:conf/eccv/ChenWZH18}.

Video is the typical sequence data. Video summarization is a sequence-to-sequence task, \emph{i.e.,} from a long sequence to short sequence \cite{zhao2020tth}. Recurrent Neural Network (RNN) is one of the effective and widely used tools for sequence modeling, which has also made tremendous success in video summarization \cite{DBLP:conf/eccv/ZhangCSG16}. Most of the recently proposed methods are developed based on the variant of RNN, \emph{i.e.,} Long Short-Term Memory (LSTM) \cite{hochreiter1997long}, and achieve the state-of-the-art performance. It mainly benefits from the great ability of LSTM in temporal dependency exploiting, which is essential for the video summarization task. However, the video structure is quite complex. It is not a smooth stream, especially for those edited videos, the multi-hop relationships occur frequently among shots \cite{wu2020dynamic}. Thus, the temporal dependency captured by LSTM is not enough to model such complex video structure. Worse still, LSTM processes the video sequentially. It requires the previous hidden states and current feature in each step, which limits its parallelization.

Transformer can remedy the above problems, which surpasses the performance of LSTM in various tasks, both in computer vision and natural language processing \cite{DBLP:conf/nips/VaswaniSPUJGKP17}. Specifically, it is designed based on the self-attention mechanism, and processes the sequence as a whole rather than step-by-step. In this case, the recursion in RNN is avoided and the parallel computation can be achieved. Moreover, it can also capture the global dependency among sequence, so that the multi-hop relationships among shots can be modeled \cite{girdhar2019video}. Inspired by this, we devote to introducing transformer to the video summarization task. 

To better utilize the transformer to summarize videos, the characteristics of video data are analyzed in this paper: 1) Video is composed of several shots recording certain activities, and each shot contains several frames varying smoothly. In this case, video data contain the two-layered hierarchical structure. 2) Videos are multimodal data. They contain both the visual and audio modalities, so as to provide viewer the sense of reality. The hierarchical and multimodal characteristics of video data are essential to the performance of video summarization.

In this paper, we develop a Hierarchical Multimodal Transformer (HMT) for the video summarization task. Specifically, a hierarchical structure for the transformer model is developed according to the video structure. Rather than processing the whole frame sequence directly, the frame sequence is separated into shots. The hierarchical transformer encodes the shot features by capturing the frame-level dependency of each shot in the first layer, and encodes the scene information by capturing the shot-level dependency in the second layer, so that the lengthy video frames are processed hierarchically. Meanwhile, the memory and computation burden are reduced. Furthermore, a multimodal fusion mechanism is developed based on the hierarchical structure, in order to jointly utilize the audio and visual information for summarizing the video. To achieve this, the transformer in the first layer is modified as two branches, \emph{i.e.,} audio branch and visual branch. The two information flows are fused in the second layer to predict the candidate summary.

Overall, the novelties and contributions of the proposed HMT are:
\begin{itemize}
	
\item The transformer model is introduced to video summarization, where the global dependency and the multi-hop relationships among shots are captured.

\item The hierarchical structure is developed for the transformer model, so that it can model the two-layered structure of video data in the summarization task.

\item The multimodal fusion mechanism is introduced to the transformer model, which can boost the performance by jointly utilizing the audio and visual information to summarize the video.
\end{itemize}

In the following, the related works in video summarization are reviewed in Section \ref{section2}. The overview of the proposed hierarchical multimodal transformer is presented in Section \ref{section3}. The experiment setup and results are discussed in Section \ref{section4}, including the ablation study and comparison with the state-of-the-art methods. The conclusions and future works are described in Section \ref{section5}.

\section{Related Works}\label{section2}

In this section, we first review the literature of video summarization, and then introduce the applications of multimodal learning in video analysis.

Traditional methods tackle video summarization as a subset selection task \cite{DBLP:journals/tip/CongLSYLL17,DBLP:conf/icip/ZhuangRHM98}. Clustering algorithms are widely adopted in earlier methods, including delaunay clustering \cite{DBLP:journals/jodl/MundurRY06}, \(k\)-means \cite{DBLP:journals/prl/AvilaLLA11} and so on. The frame sequence is allocated into several clusters. The cluster centers are determined as representatives. Dictionary learning is another tool for subset selection. The regularization term defined by \(l_0\) norm, \(l_{0,1}\) norm and rank norm are usually integrated into the dictionary learning process to guarantee the sparsity of selected key-frames or key-shots \cite{DBLP:conf/icmcs/MeiGWHHF14,DBLP:journals/tmm/CongYL12}. Other constraints are also employed to model the priors of the video data, such as local similarity and smoothness \cite{DBLP:journals/tmm/LuWMGF14,ma2020video}. Clustering algorithms and dictionary learning select summary based on low-level video features. To employ high-level features, different score functions are designed to rank video shots \cite{DBLP:journals/tip/LiZL17,tschiatschek2014learning}. They are developed based on the object-level information (size, location, occurrence frequency, \emph{etc.}), smoothness of video storyline, distribution of key-shots, diversity of the key-shot set, and so on  \cite{DBLP:journals/ijcv/LeeG15,DBLP:conf/cvpr/LeeGG12}. By utilizing high-level features, the performances are improved. 

In recent years, RNN-based methods are rising rapidly with the development of deep learning. The plain LSTM is firstly utilized in \cite{DBLP:conf/eccv/ZhangCSG16} to encode the temporal dependency among frame sequence. The frame-wise probability is predicted by the multi-layer perception. Compared to traditional methods, better results are obtained by taking advantage of the temporal dependency captured by RNN. To adapt to the video data, the hierarchical structure is developed for RNN to extend the long sequence modeling ability \cite{zhao2020tth}. The hierarchical transformer in our work is inspired from it. The boundary-aware RNN is also proposed to jointly detect shot boundaries and select key-shots \cite{DBLP:conf/cvpr/ZhaoLL18}. To provide more priors for the summarizer, the generative adversarial network is utilized in \cite{DBLP:conf/cvpr/MahasseniLT17}, and the discriminative loss is used for optimization. The score functions are borrowed from traditional methods as the reward, in order to conduct reinforcement learning to optimize the summary generator \cite{DBLP:conf/aaai/ZhouQX18}. Dual learning is also adopted in a similar way, which summarizes the video under the supervision of  the reverse task, \emph{i.e.,} video reconstruction \cite{zhao2019property}. With the help of discriminator, reinforcement learning and dual learning, the video can be summarized unsupervisedly. Although great success is achieved by RNN-based methods, they can only capture the temporal dependencies, while the global dependencies are neglected. In this case, the multi-hop relationships among shots cannot be captured \cite{wu2020dynamic}. 

Researchers have realized the limitations of RNN in the video summarization task. They try to introduce attention models 
to capture the global dependencies \cite{wang2020query,wang2020transfoming}. On the one hand, the attention model is integrated into the RNN model. Specifically, the attentive encoder-decoder network is proposed in \cite{ji2019video}. The decoder is composed with two bidirectional LSTMs equipped with the attention model. An attentive and distribution consistent method is proposed in \cite{ji2020deep}, where the self-attention is applied on the encoder, and the additive and multiplicative attention is exploited on the decoder. The whole network is optimized jointly by the regression loss and distribution loss. On the other hand, the pure attention model is utilized for video summarization, which means RNN is abandoned. The global diverse attention is explored in \cite{li2021exploring} by capturing the pairwise relation among video frames. It can compute the representativeness of each frame to the whole video content and lead to diverse attention. The user's preference are considered in \cite{xiao2020convolutional} by developing a query-focused video summarization framework, where both self-attention and query-aware attention are employed. 

Finally, we want to emphasize the development of multimodal learning in video analysis. Video captioning, sounding object localization, audiovisual representation are typical tasks of multimodal learning, where the audio, visual and textual modalities are utilized for video analysis \cite{wei2019neural,tsiami2020stavis,scholes2020interrelationship}. A temporal fusion mechanism is developed in \cite{DBLP:conf/icmi/VielzeufPJ17} to fuse the audio and visual data for emotion classification. The multimodal latent topics are mined in \cite{DBLP:conf/mm/ChenCJH17} to generate video captions. Similar to the video summarization task, an automatic curation method of sports highlights is conducted by combining multimodal excitement features \cite{DBLP:journals/tmm/MerlerMJNHKXDSF19}. Similarly, the audiovisual features are jointly utilized to summarize baseball videos in \cite{lee2020hierarchical}. Moreover, a neural multimodal cooperative learning framework is developed in \cite{DBLP:journals/tip/WeiWGNLC20} for video understanding tasks. Generally, most above multimodal works show superiority than those methods just utilizing visual features.

\begin{figure*}[t]
	\centering
	\includegraphics[width=0.98\textwidth]{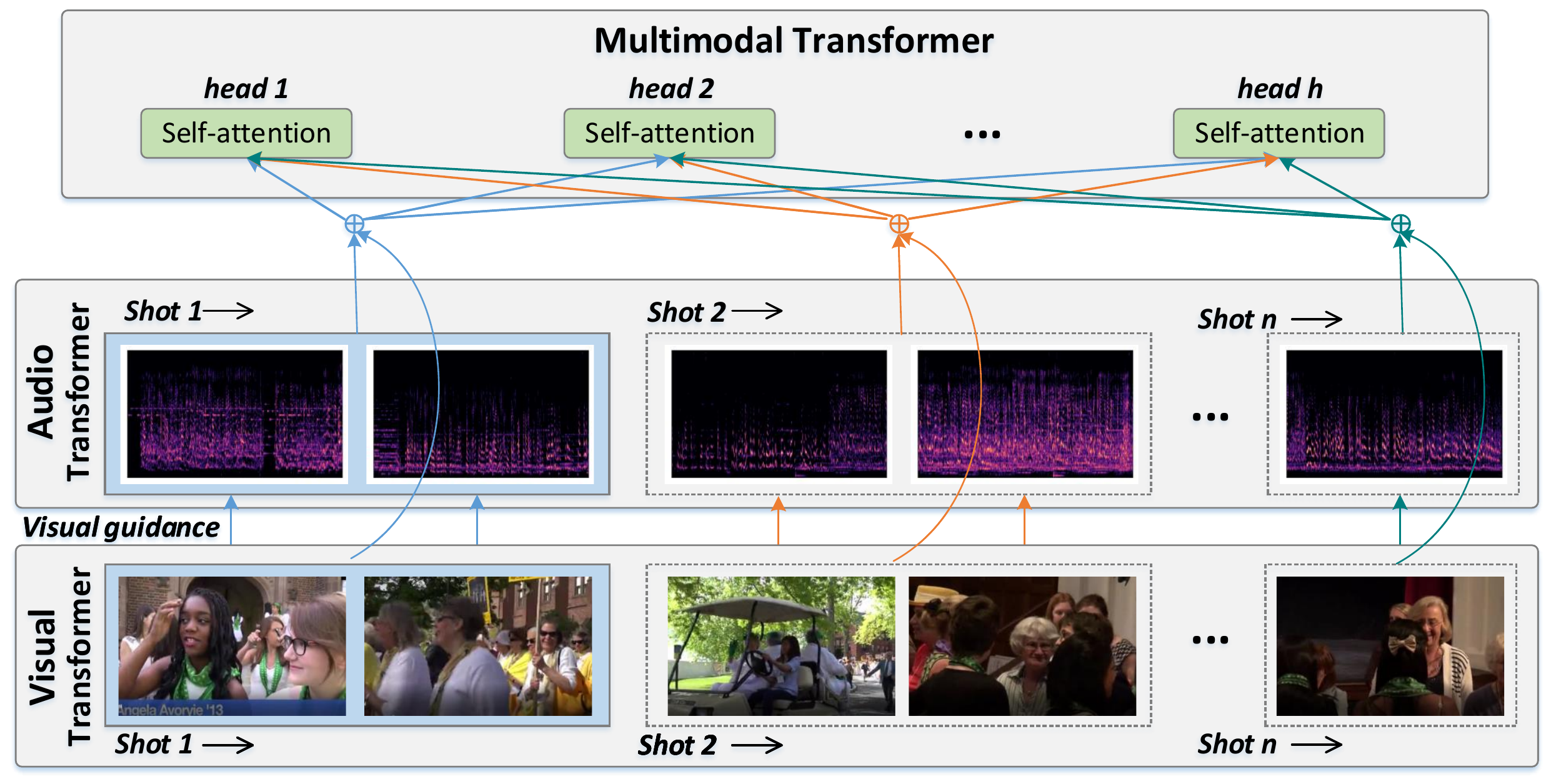}
	\caption{The full architecture of the proposed hierarchical multimodal transformer, where the first layer is composed of a visual transformer and an audio transformer, and the second layer is a multimodal transformer taken the fused audiovisual information as input. Note that \emph{Visual guidance} means the audio feature is encoded under the guidance of visual feature. The lines in different color denote the information from different video shots. }\label{fig2}
\end{figure*}

\section{The Proposed Method}\label{section3}

In this paper, a Hierarchical Multimodal Transformer (HMT) is proposed for the video summarization task. As depicted in Figure \ref{fig2}, the transformer is utilized to encode the global dependency in the video stream. The hierarchical structure is developed to model the video in frame-level and shot-level successively. The multimodal fusion mechanism is presented to integrate the audio and visual information for video summarization. In the following subsections, each component is elaborated in detail.

\subsection{The Plain Transformer Encoder}\label{section3.1}

Transformer is originally proposed for the machine translation task, which translates the sentence from source language to target language. It contains an encoder and a decoder. They share similar architectures with a multi-head self-attention layer and a feed-forward network. In this work, only the encoder is utilized for video summarization.

Given the input sequence \(X=\left[ {{x_1},{x_2}, \ldots ,{x_n}} \right]\), each element is transformed into three kinds of vectors, 
\begin{equation}Q = {W^Q}X,\;K = {W^K}X,\;V = {W^V}X,\end{equation}
where \(Q,\;K,\;V \in {\mathbb{R}^{d\times n}}\) denote the query, key and value matrices, respectively. \(W^Q\), \(W^K\), \(W^V\) are the linear mapping parameters. Then, the self-attention is conducted by 
\begin{equation}SA = softmax\left( {\frac{{Q \cdot {K^\top}}}{{\sqrt d }}} \right) \cdot V,\end{equation}
where \(Q \cdot {K^\top}\) computes the pairwise relationship (similarity) in the sequence. The softmax function transforms the similarity to attention weights.

To jointly model dependencies from different representation subspaces at different positions, the multi-head attention is adopted. The above self-attention can be viewed as single-head attention. The multi-head attention is achieved by conducting self-attention in different representation subspaces. By utilizing different linear mapping matrices, \(W_i^Q\), \(W_i^K\), \(W_i^V\), \(i = 1,2, \ldots ,h\), the representations of query, key and value can be obtained as 
\(\left\{ {{Q_i},{K_i},{V_i}} \right\}_{i = 1}^h\). \(h\) is a hyper-parameter denoting the number of heads in the multi-head attention. The self-attention can be conducted in different subspaces, \emph{i.e.,}
\begin{equation}S{A_i} = softmax\left( {\frac{{{Q_i} \cdot K_i^ \top}}{{\sqrt d }}} \right) \cdot {V_i},\quad i = 1,2, \ldots ,h,\end{equation}
then, the multi-head attention is formulated as 
\begin{equation}MHA = \left[ {S{A_1},S{A_2}, \ldots ,S{A_h}} \right] \cdot {W^o},\end{equation}
where \(W^o\) is the training parameter. 

After that, a feed-forward network and a layer-norm operation are stitched after the multi-head attention layer. The frame features are further encoded as
\begin{equation} E = LN\left({W^2} ReLU \left( {{W^1}MHA} \right)\right),\end{equation}
where \(E\) is the encoded feature matrix. \(W^1\) and \(W^2\) are the training parameters. \(ReLU\) is the activation function. \(LN\) denotes the layer-norm operation.
\subsection{Hierarchical Multimodal Transformer}

According to \cite{DBLP:conf/eccv/ZhangGS18,DBLP:conf/cvpr/PanXYWZ16}, videos are naturally two-layered data, \emph{i.e.,} frame-shot-video. Inspired by this, the hierarchical structure is fixed as two layers, where the first layer processes the frames in each shot, and the second layer processes all the shots in the video.

The first step for conducting the hierarchical structure is to separate the frame sequence into shots. Kernel-based Temporal Segmentation (KTS) \cite{DBLP:conf/eccv/PotapovDHS14} is the most widely used shot boundary detection method. Following existing protocols, it is also adopted in our method for fairness. 
Given the frame sequence \(\left[ {{f_1},{f_2}, \ldots ,{f_n}} \right]\), it is separated into shots by KTS, and the subsequences are obtained as \(\left[ {{f_{{b_i}}},{f_{{b_i} + 1}}, \ldots ,{f_{{b_{i + 1}}-1}}} \right],\; i = 0,1, \ldots ,m - 1\), where \(m\) denotes the number of shots in the video, \(\left\{ {{b_i}} \right\}_{i = 0}^m\) are the shot boundaries. Note that \({b_0} = 1\) and \({b_m} = n\) stands for the start frame and end frame of the video. The audio data is segmented temporally according to the same boundaries.

\subsubsection{Frame-level Transformer}

 To encode the visual and audio information jointly, the first layer of the hierarchical transformer is modified as two branches, \emph{i.e.}, the visual branch and audio branch. 

In the visual branch, the frames are encoded by a plain transformer for each shot. Without loss of generality, the encoding of the \(i\)-th shot is formulated as 
\begin{equation}E_i^1 = FTrans\left( {\left[ {{f_{{b_{i-1}}}},{f_{{b_{i-1}} + 1}}, \ldots ,{f_{{b_{i}}}}} \right]} \right),\end{equation}
where \(FTrans\) denotes all the computations in the frame-level transformer, as depicted in Section \ref{section3.1}. \(E_i^1\) is the encoded features of frames in shot \(i\). To obtain the visual representation vector of the shot, the mean pooling operation is carried out on \(E_i^1\),
\begin{equation}{s_i^v} = MP\left( {E_i^1} \right),\end{equation}
where \(MP\) stands for the mean pooling operation. In this case, the visual representations for each shot can be obtained as 
\({\left[ {{s_1^v},{s_2^v}, \ldots ,{s_m^v}} \right]}\).

There are inconsistency between audio and visual information, \emph{e.g.}, the sounding objects are not recorded in the video frames. To encode the visual-related audio information, in the audio branch, the audio feature is encoded under the guidance of visual feature. It is formulated as
\begin{equation}S{A^a_i} = softmax\left( {\frac{{{Q_i^v} \cdot K_i^{a \top}}}{{\sqrt d }}} \right) \cdot {V_i^a},\quad i = 1,2, \ldots ,h,\end{equation}
where \(Q^v\) is the query matrix of the visual feature. \(K^a\) and \(V^a\) denote the key and value matrices of the audio feature. Under the guidance of visual features, Eqn. (8) can explore the consistency between audio and visual features, so as to reduce the interference caused by the inconsistency. With the exception of self-attention, the audio and visual branches in the first layer share the same architecture. The final computed audio feature of each shot is denoted as \({\left[ {{s_1^a},{s_2^a}, \ldots ,{s_m^a}} \right]}\).

To jointly utilize the visual and audio information for summarization,  the two kinds of features are concatenated together to form the final shot representation, \emph{i.e.},
\begin{equation}{s_i} = [s_i^v;s_i^a],\quad i = 1, \ldots ,m,\end{equation}
where \([ \cdot ; \cdot ]\) stands for the concatenation of two vectors. Taking the audio-visual features as input, the multimodal fusion process is carried out in the second layer.

\subsubsection{Shot-level Transformer}

The shot representations are input to the second layer, in order to capture the global dependencies among video shots. Specifically, the transformer in the second layer share similar architectures with the one in the first layer. The computation is formulated as 
\begin{equation}E^2 = STrans\left( {{\left[ {{s_1},{s_2}, \ldots ,{s_m}} \right]}} \right),\end{equation}
where \(STrans\) denotes the computations of the shot-level transformer, and \(E^2\) is the encoded shot features.

Finally, the video features are encoded in the hierarchical transformer. The probability of each shot to be selected into the summary is computed as
\begin{equation}P = softmax\left( {{W^P}{E^2}} \right).\end{equation}

\subsection{Optimization}
In this paper, the Mean Square Error (MSE) is employed for the optimization of the hierarchical multomodal transformer, \emph{i.e.,}
\begin{equation}MSE = \frac{1}{n}\left\| {\tilde P - G} \right\|_2^2,\end{equation}
where \(\tilde P\) and \(G\) are the frame-level probability vector predicted by the proposed method and annotated by human beings. Note that \(\tilde P\) is extended from the shot-level probability \(P\) computed by Eqn. (11), where each frame is assigned with the probability of corresponding shot.


\section{Experiments}\label{section4}
The proposed HMT is evaluated on two datasets, SumMe \cite{DBLP:conf/eccv/GygliGRG14} and TVsum \cite{song2015tvsum}. The results are compared with several state-of-the-art methods to verify its superiority. Besides, the ablation study is conducted to show the improvements of each component of HMT. 

\begin{table*}[htp]
	\centering\small
	\caption{The statistics of datasets.}\label{Table1}
	\renewcommand\arraystretch{1}\footnotesize
	
	\begin{tabularx}{\textwidth}{X<{\centering}|X<{\centering}|X<{\centering}}
		\toprule\hline

		Name &Number of videos &Average duration   \\\hline
		SumMe \cite{DBLP:conf/eccv/GygliGRG14}  &25 &1-6min \\
		TVsum \cite{song2015tvsum}  &50 &2-10min \\
		YouTube \cite{DBLP:journals/prl/AvilaLLA11}  &39 &1-10min \\
		OVP \cite{DBLP:journals/prl/AvilaLLA11}  &50 &1-4min \\\hline\bottomrule 
	\end{tabularx}
	
\end{table*}

\subsection{Setup} 
\subsubsection{Dataset Introduction} 

Four datasets are employed in this paper, including SumMe \cite{DBLP:conf/eccv/GygliGRG14}, TVsum \cite{song2015tvsum}, YouTube \cite{DBLP:journals/prl/AvilaLLA11} and OVP \cite{DBLP:journals/prl/AvilaLLA11}. They are all collected from the Internet and share similar video topics, such as traveling, cooking, news, \emph{etc.}
Their statistics are displayed in Table \ref{Table1}. Following existing protocols \cite{DBLP:conf/eccv/ZhangCSG16,DBLP:conf/eccv/PotapovDHS14,ma2020video}, SumMe and TVsum are utilized for evaluation. The summarization rate is fixed as 15\%, where the key-shots are selected by dynamic programming \cite{DBLP:conf/aaai/ZhouQX18} as the knapsack problem. The training/test rate is 80\% and 20\%. The training set also plays the role of validation. To make the results more convincing, 5 random splits of training/test set are evaluated, and the average results are reported in this paper. YouTube and OVP are used in the augmented setting to augment the training data.

Both the visual features and audio features are taken as the input of the proposed HMT. The 1024-dim feature in the pool-5 layer of GoogLeNet \cite{DBLP:conf/cvpr/SzegedyLJSRAEVR15} is adopted as the visual feature of frames. The 128-dim feature of VGGish \cite{hershey2017cnn} is employed as the audio feature. To align the visual and audio feature temporally, the visual feature is extracted for 2 frames per second. The audio data is segmented with the duration of 1 second and the overlap of 0.5 second.

\subsubsection{ Evaluation Metrics}

Two kinds of evaluation metrics are employed in the experiment. F-measure is the most popular metric in video summarization. It measures the summary quality by the temporal overlap between the summaries generated automatically and annotated by human. Besides, the rank-based evaluation metrics, Kendall's \(\tau\) and Spearman's \(\rho\) \cite{otani2019rethinking} are adopted. They measure the similarity between predicted probabilities and annotated importance scores, which provide a more fine-grained evaluation of the summary quality. Practically, there are multiple annotated summaries and important scores for each video. The pairwise evaluation is conducted for each annotation. Following existing protocols \cite{DBLP:conf/eccv/ZhangCSG16,DBLP:conf/eccv/RochanYW18,ji2019video,DBLP:conf/eccv/ZhangGS18}, the average results are reported for the two kinds of metrics on each dataset, except for the F-measure on SumMe whose maximum results are reported.

\subsubsection{Optimization Details}
The proposed HMT is implemented with PyTorch 1.6 on Python 3.7. Specifically, HMT is optimized by minimizing the MSE loss via the Adam optimizer, where the learning rate is initialized as 1e-4, with the decay rate 0.1. The dimensionality of the hidden state in the transformer is fixed as 1024. The number of heads in the multi-head attention is 4. The number of layers of the transformer is fixed as 1. The dimensionality in each subspace is 64.

\subsection{Comparison with Baselines}
The proposed HMT can be split into three folds: 1) The transformer can capture the global dependency. 2) The hierarchical structure is more suitable for the two-layered video structure. 3) The multimodal fusion mechanism can integrate the visual and audio information. To verify the effectiveness of the above three parts, several baselines are developed according to HMT, which are depicted as follows:

\begin{table*}[tp]
	\centering
	\caption{Comparisons between HMT and baselines (mean(std) of F-measure).}\label{Table2}
	\renewcommand\arraystretch{1.0}\footnotesize
	
	\begin{tabularx}{\textwidth}{p{6cm}<{\centering}|X<{\centering}|X<{\centering}}
		\toprule\hline
		
		Baselines &SumMe&TVsum\\ \hline
		Audio LSTM&0.354(0.062)&0.539(0.025)\\
		Visual LSTM&0.397(0.034)&0.556(0.021)\\
		Audio Transformer &0.378(0.051) &0.548(0.023)  \\
		Visual Transformer &0.403(0.023) &0.569(0.020)  \\
		Two-stream Transformer &0.394(0.027) &0.562(0.018)  \\
		Multimodal Transformer &0.415(0.022) &0.579(0.021) \\
		Hierarchical Transformer &0.434(0.031) &0.591(0.015)  \\
		Hierarchical Multimodal Transformer&0.441(0.021)&0.601(0.016) \\
		\hline\bottomrule
	\end{tabularx}
	
\end{table*}

\begin{itemize}
\item Audio LSTM and Visual LSTM: The plain LSTM is utilized to predict the summary. They take the audio feature and visual feature as the input, respectively.

\item Audio Transformer and Visual Transformer: They encode the audio feature and visual feature by the plain transformer model, respectively.

\item Two-stream Transformer: Two transformers are employed to encode the visual and audio features individually, where the visual feature provides no guidance to the audio transformer. The probability is predicted by simply concatenating the encoded visual and audio features.

\item Multimodal Transformer: The two-branch transformer in the first layer of HMT is utilized to encode the visual and audio features. It should be emphasized that the audio feature is encoded under the guidance of visual feature.

\item Hierarchical Transformer: The proposed hierarchical structure is utilized to encode the video and predict the summary just with the visual feature.

\item Hierarchical Multimodal Transformer: The full model of the proposed method.
\end{itemize}

The comparisons of different baselines are shown in Table \ref{Table2}. It can be observed that the plain LSTMs get satisfactory results by taking audio or visual information as input, which shows the contributions of both audio and visual information to video summarization. The large standard deviation of audio LSTM is mainly because two videos in SumMe do not contain audio data. Transformers surpass plain LSTMs by modeling the frame or audio sequence as a whole and capturing global dependency. The improvements meet our expectations. It also demonstrates the superiority of transformer than LSTM in video summarization. Surprisingly, the two-stream transformer performs slightly worse than the plain transformer, but the multimodal transformer performs much better. It is because there are modal gaps between the audio and visual information. The interference caused by concatenating them directly without exploiting the consistency may surpass the gains from them. As a result, the performance is decreased. On the opposite side, the better performance of multimodal transformer indicates the effectiveness of fusing visual and audio information for video summarization.

The hierarchical transformer outperforms the plain models, including LSTM, transformer, two-stream transformer and multimodal transformer. It mainly benefits from the hierarchical structure. It is more suitable for the video data, since frames form shot, shots form the video. The two-layered hierarchical transformer matches the key-shot based video summarization task well. Besides, the full method, hierarchical multimodal transformer, can further improve the performance by fusing the visual and audio information.

Furthermore, the parameter analysis is conducted on the backbone transformer, including the number of layers and heads. The results are depicted in Table \ref{Table3add} and \ref{Table3add2}. It can be observed that the proposed HMT performs stably with the variation of heads and layers of the backbone transformer, which shows the robustness of HMT. Carefully, we can see that HMT performs the best when the number of heads and layers are fixed as 4 and 1, respectively. In this case, they are fixed in all the following experiments. 
\begin{table*}[tp]
	\centering
	\caption{Parameter analysis on transformer about the number of heads.}\label{Table3add}
	\renewcommand\arraystretch{1.0}\footnotesize
	
	\begin{tabularx}{\textwidth}{X<{\centering}|X<{\centering}|X<{\centering}}
		\toprule\hline
		
		heads &SumMe&TVsum\\ \hline
		1 &0.414&0.579\\
		2 &0.425 &0.585  \\
		4 &0.441 &0.601 \\
		8 &0.439 &0.597 \\
		16 &0.439 &0.596  \\
		\hline\bottomrule
	\end{tabularx}
	
\end{table*}

\begin{table*}[tp]
	\centering
	\caption{Parameter analysis on transformer about the number of layers.}\label{Table3add2}
	\renewcommand\arraystretch{1.0}\footnotesize
	
	\begin{tabularx}{\textwidth}{X<{\centering}|X<{\centering}|X<{\centering}}
		\toprule\hline
		
		layers &SumMe&TVsum\\ \hline
		1 &0.441&0.601\\
		2 &0.433 &0.592  \\
		3 &0.414 &0.578 \\
		\hline\bottomrule
	\end{tabularx}
	
\end{table*}

\subsection{Comparison with State-of-the-art Methods}

To verify the superiority of the proposed HMT in video summarization, the comparisons are in three aspects, \emph{i.e.,} traditional methods, RNN-based methods and attention-based methods. 

\begin{table*}[tp]
	\centering
	\caption{Comparisons between HMT and traditional methods (F-measure).}\label{Table3}
	\renewcommand\arraystretch{1.0}\footnotesize
	
	\begin{tabularx}{\textwidth}{X<{\centering}|X<{\centering}|X<{\centering}}
		\toprule\hline
		
		Datasets &SumMe&TVsum\\ \hline
		\(k\)-medoids \cite{hadi2006video} &0.334&0.288\\
		Delauny \cite{DBLP:journals/jodl/MundurRY06}  &0.315 &0.394  \\
		VSUMM \cite{DBLP:journals/prl/AvilaLLA11} &0.335 &0.391 \\
		SALF \cite{DBLP:conf/cvpr/ElhamifarSV12} &0.378 &0.420  \\
		LiveLight \cite{DBLP:conf/cvpr/ZhaoX14a} &0.384&0.477 \\
		Block Sparse \cite{ma2020video} &0.401&0.526 \\
		CSUV \cite{DBLP:conf/eccv/GygliGRG14}&0.393	&0.532	\\
		LSMO \cite{gygli2015video} &0.403	&\underline{0.568}	\\
		Summary Transfer \cite{DBLP:conf/cvpr/ZhangCSG16} &\underline{0.409}	&--	\\
		HMT &\textbf{0.441}	&\textbf{0.601}\\
		\hline\bottomrule
	\end{tabularx}
	
\end{table*}

The comparisons of HMT and traditional methods are depicted in Table \ref{Table3}. The clustering appproaches (\emph{i.e.,} \(k\)-medoids, Delauny, VSUMM) and dictionary learning methods (\emph{i.e.,} SALF, LiveLight and Block Sparse) are developed based on low-level features. Generally, dictionary learning methods select key-shots via reconstructing the video from summary. It can be viewed as the global dependency modeling. However, the clustering methods focus more on the local similarity among frames. That is why dictionary learning methods perform better than clustering methods. CSUV, LSMO and Summary Transfer are developed based on high-level features. They develop various models to measure the property of generated summaries. Obviously, they outperform those methods with low-level features. The proposed HMT is a deep learning based method. It performs much better than traditional methods compared in Table \ref{Table3}, which shows the great learning ability.

\begin{table*}[tp]
	\centering
	\caption{Comparisons between HMT and RNN-based methods.}\label{Table4}
	\renewcommand\arraystretch{1.0}\footnotesize
	
	\begin{tabularx}{\textwidth}{p{4cm}<{\centering}|X<{\centering}X<{\centering}|X<{\centering}X<{\centering}}
		\toprule
		\hline
		Datasets &\multicolumn{2}{c}{SumMe}&\multicolumn{2}{c}{TVsum}\\\hline
		Organizations &Canonical &Augmented&Canonical  &Augmented  \\\hline
		vsLSTM \cite{DBLP:conf/eccv/ZhangCSG16} &0.376(0.022) &0.416&0.542(0.016) & 0.579   \\
		dppLSTM \cite{DBLP:conf/eccv/ZhangCSG16} &0.386(0.023) &0.429&0.547(0.018) &0.596   \\
		SUM-GAN \cite{DBLP:conf/cvpr/MahasseniLT17}&0.387 &0.417 &0.508&0.589 \\
		
		SUM-GAN\(_{sup}\) \cite{DBLP:conf/cvpr/MahasseniLT17} &0.417&0.436 &0.563&\underline{0.612}\\
		H-RNN \cite{DBLP:conf/mm/ZhaoLL17}&0.421	&0.438 &0.579 &\textbf{0.619}	\\
		HSA-RNN \cite{DBLP:conf/cvpr/ZhaoLL18}&0.423(0.026) &0.421	&\underline{0.587}(0.017) &0.598	\\	
		DR-DSN \cite{DBLP:conf/aaai/ZhouQX18}&0.414(0.036) &0.428	&0.576(0.022) &0.584	\\
		DR-DSN\(_{sup}\) \cite{DBLP:conf/aaai/ZhouQX18}&0.421(0.041) &0.439	&0.581(0.024) &0.598	\\	
		
		SMIL \cite{lei2018action}&0.412 &--	&0.513&--	\\
		WS-HRL \cite{DBLP:conf/mmasia/ChenTWY19} & \underline{0.436}& \underline{0.445} &0.584&0.585\\
		HMT &\textbf{0.441}(0.021)	&\textbf{0.448}	&\textbf{0.601}(0.016)	&0.603	\\
		\hline\bottomrule
	\end{tabularx}
	
\end{table*}

The comparisons of HMT and RNN-based methods are depicted in Table \ref{Table4}. The results of both the canonical and augmented settings are presented in Table \ref{Table4}. In the canonical setting, the training/test split is 80\% and 20\% on SumMe and TVsum, and they are trained separately. In the augmented setting, the training set is formed by 80\%*SumMe+YouTube+OVP+TVsum, and the rest 20\%*SumMe is used for testing. Similarly, 80\%*TVsum+YouTube+OVP+SumMe is utilized as the training set, and the rest 20\%*TVsum is used for testing. It can be observed from Table \ref{Table4} that the performances of most methods are promoted by the augmentation of training data.

RNN-based methods process the video data with a plain model (\emph{i.e.,} vsLSTM, dppLSTM) or hierarchical model (\emph{i.e.,} H-RNN, HSA-RNN). It can be observed that the hierarchical model performs much better than the plain models, which indicates the superiority of hierarchical model. The hierarchical structure of transformer is also inspired from it. To make up for the deficiency of plain models, SUM-GAN utilizes the generative adversarial network to boost the performance. DR-DSN conducts a reinforcement learning framework to reward the summary generator with summary properties. The reinforcement learning is also adopted in SMIL and WS-HRL, and better performances are achieved. However, they are all developed based on LSTM, where only the temporal dependencies are captured. Fortunately, the proposed HMT can better capture the global dependency, which is more suitable for the video summarization task. Besides, all the compared methods just utilize visual features for summarization, while the multimodal fusion mechanism in HMT integrates the visual and audio information together. The multimodal information also contributes to the performance.

\begin{table*}[tp]
	\centering
	\caption{Comparisons between HMT and attention-based methods.}\label{Table5}
	\renewcommand\arraystretch{1.0}\footnotesize
	
	\begin{tabularx}{\textwidth}{p{4cm}<{\centering}|X<{\centering}X<{\centering}|X<{\centering}X<{\centering}}
		\toprule
		\hline
		Datasets &\multicolumn{2}{c}{SumMe}&\multicolumn{2}{c}{TVsum}\\\hline
		Organizations &Canonical &Augmented&Canonical  &Augmented  \\\hline
		vsLSTM-att\cite{DBLP:conf/mmm/CasasK19} &0.432(0.028) &-- &--&--\\
		dppLSTM-att\cite{DBLP:conf/mmm/CasasK19} &0.438(0.022) &--&0.539(0.051)&--\\
		A-AVS \cite{ji2019video}&0.439 &\underline{0.446}	&\underline{0.594} &\textbf{0.608}	\\
		VASNet \cite{fajtl2018summarizing}&0.424 &0.425	&0.589 &0.585	\\	
		SASUM \cite{DBLP:conf/aaai/WeiNYYYY18}&0.406 &--	&0.539&--	\\
		SASUM\(_{sup}\) \cite{DBLP:conf/aaai/WeiNYYYY18}&\textbf{0.453}&--	&0.582&--	\\
		
		HMT &\underline{0.441}(0.021)&\textbf{	0.448}&\textbf{0.601}(0.016)&\underline{0.603}\\
		\hline\bottomrule
	\end{tabularx}
	
\end{table*}

The comparisons of HMT and attention-based methods are depicted in Table \ref{Table5}. Note that there are several blanks. It is because the results are not reported in those papers and the source codes are not available. vsLSTM-att and dppLSTM-att are modified from pure RNN-based methods by adding the attention model in the encoder. By comparing with the results of vsLSTM and dppLSTM in Table \ref{Table4}, we can clearly see the improvements of the attention model. It meets our motivation to exploit the global dependencies for video summarization. VASNet\footnote{For fair comparison, the results of VASNet are reproduced by keeping its experimental settings the same with ours.} and A-AVS are pure attention-based methods designed based on the self-attention model. Carefully, we can see that the performance of VASNet goes down from the canonical setting to the augmented setting. It is mainly because that the interference cause by the difference among training set exceeds the benefit of training data augmentation. In fact, the self-attention model is also adopted in the proposed method. The proposed transformer-based method exceeds it by taking advantage of the multi-head setting. SASUM is a caption-aided multimodal video summarization method. It summarizes the video under the supervision of video captions. Although better performance is achieved by SASUM\(_{sup}\) on SumMe, the captions are hardly available in most occasions, which limits its applications. Fortunately, the visual and audio data are aligned naturally in videos. The proposed HMT is more applicable to the video summarization task.

\subsection{Rank-based Evaluation}
\begin{table*}[tp]
	\centering
	\caption{Comparisons based on the metrics of Kendall's \(\tau\) and Spearman's \(\rho\).}\label{Table7}
	\renewcommand\arraystretch{1.0}\footnotesize
	
	\begin{tabularx}{\textwidth}{p{3cm}<{\centering}|X<{\centering}X<{\centering}|X<{\centering}X<{\centering}}
		\toprule
		\hline
		Datasets &\multicolumn{2}{c}{SumMe}&\multicolumn{2}{c}{TVsum}\\ \hline
		Metrics &Kendall's \(\tau\)  & Spearman's \(\rho\) &Kendall's \(\tau\)  & Spearman's \(\rho\)  \\ \hline

		dppLSTM \cite{DBLP:conf/eccv/ZhangCSG16}  &-- &--&0.042&0.055   \\
		
		DR-DSN \cite{DBLP:conf/aaai/ZhouQX18} &0.047&0.048&0.020&0.026 \\
		HSA-RNN \cite{DBLP:conf/cvpr/ZhaoLL18}&0.064&\underline{0.066}&\underline{0.082}&0.088\\
		vsLSTM-att\cite{DBLP:conf/mmm/CasasK19} &0.062 &0.063 &0.074&0.083\\
        dppLSTM-att\cite{DBLP:conf/mmm/CasasK19} &\underline{0.066} &0.065&0.076&0.081\\
        VASNet \cite{fajtl2018summarizing}&0.054 &0.058	&\underline{0.082} &0.088	\\
		
		WS-HRL \cite{DBLP:conf/mmasia/ChenTWY19} & --& -- &0.078&\textbf{0.116}\\
		Transformer & 0.060& 0.062 &0.081&0.095\\
		HMT &\textbf{0.079}	&\textbf{0.080}	&\textbf{0.096}	&\underline{0.107}\\\hline
		Random selection  &0.000 &0.000&0.000 &0.000  \\
		Human &0.205&0.213&0.177&0.204\\
		\hline\bottomrule
	\end{tabularx}
	
\end{table*}

F-measure evaluates the summary quality by measuring the overlap between predicted summary and reference summary. Different from F-measure, rank-based metrics evaluate the performance by computing the correlation between the probability curves predicted automatically and annotated by human. Therefore, the two kinds of metrics can cooperate with each other so as to provide comprehensive evaluation of the summary quality.

Two rank-based metrics are employed in this paper, including Kendall's \(\tau\)  and Spearman's \(\rho\). The results are shown in Table \ref{Table7}. The metric values are positively correlated to the performance. It can be observed that the performances of random selection and human annotation are the lowest and highest, respectively, which meet the expectations. HMT outperforms most of the compared methods. Particularly, HMT performs much better than the plain transformer. Overall, the results in Table \ref{Table7} demonstrate the advantages of the proposed HMT: 1) Transformer can model the global dependencies among frames, which is essential for video summarization. 2) The hierarchical structure is more suitable for the two-layered video structure, and can extend the ability of plain transformer in long sequence modeling. 3) The multimodal fusion mechanism can further promote the performance by integrating visual and audio information together. 

\begin{figure*}[t]
	\centering
	\includegraphics[width=0.68\textwidth]{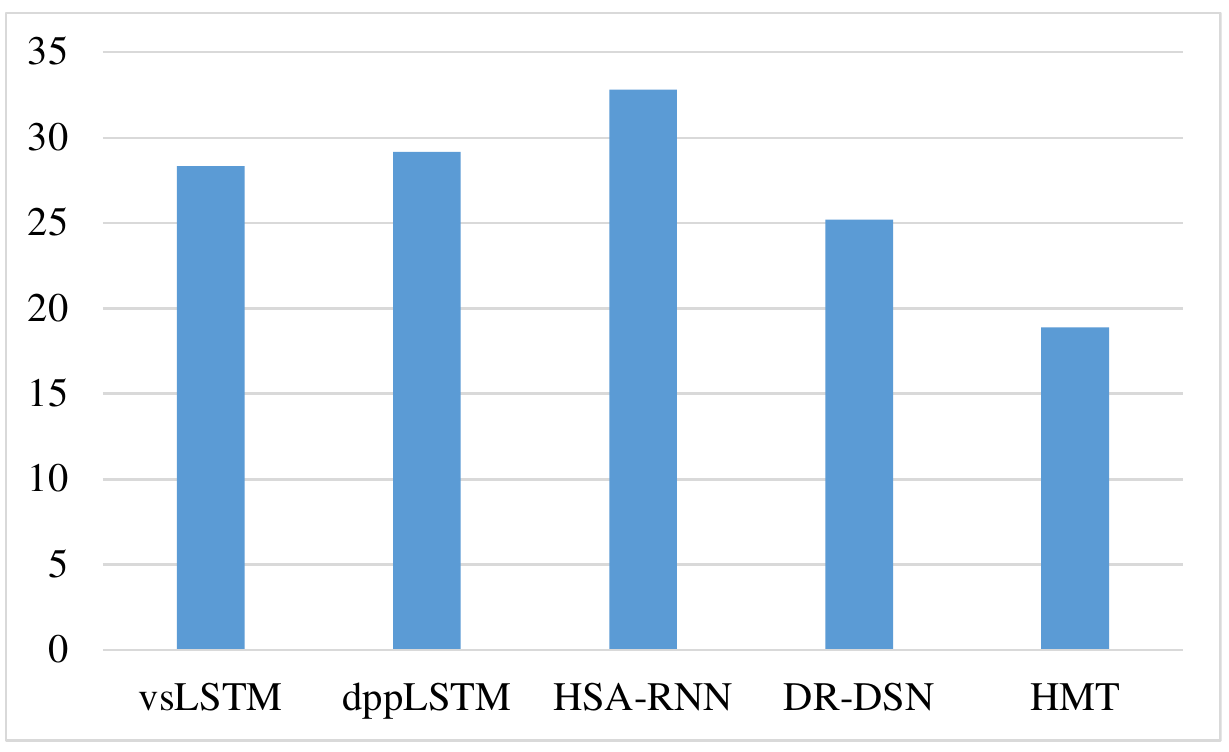}
	\caption{Computation time (ms) of different methods. }\label{fig3}
\end{figure*}

\subsection{Runtime Comparison and Visualization}

The computation time of different methods are compared in Figure \ref{fig3}. For fairness, all the compared methods and the proposed HMT are operated on Nvidia GTX 1080Ti. It records the test time from feature input to summary generated, which is further normalized by the video duration (in minutes) to reveal the influence of sequence length. We can see that the proposed HMT is more efficient than the compared RNN-based methods, even though HMT contains a hierarchical structure and a two-branch structure in the first layer. It demonstrates the efficiency of transformer than RNN, and the effectiveness of transformer in parallel computation.

\begin{figure*}[t]
	\centering
	\includegraphics[width=0.98\textwidth]{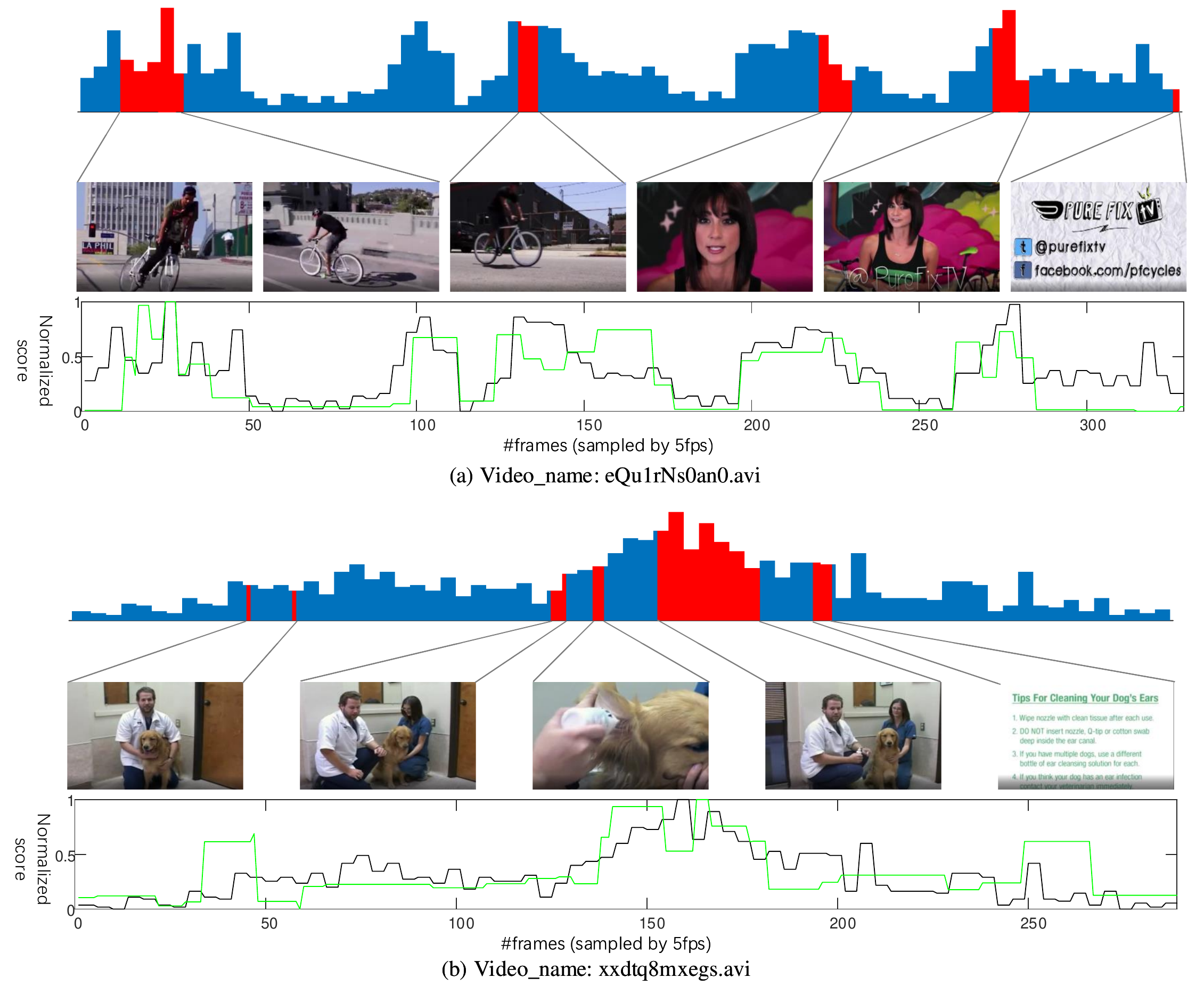}
	\caption{Example summaries generated by HMT. The green curves are predicted by HMT. The black curves are ground truth. The red histograms denote the selected key-shots.}\label{fig4}
\end{figure*}
To provide a comprehensive understanding of the performance, some video summaries and probability curves predicted by HMT are visualized in Figure \ref{fig4}. We can see that the predicted curves fit well with the human-annotated importance scores. Most of the shots with higher scores are selected into the summary, which shows the effectiveness of HMT in summarizing the video. Finally, we want to emphasize that the second and fourth shots in Figure \ref{fig4}(b) are quite similar visually, but they display different audio information. It is mainly benefit from our multimodal strategy in the proposed hierarchical multimodal transformer, so that it can recognize the importance and difference of these two shots and select them together. Beside, the very last selected key-shot in Figure \ref{fig4}(b) is a failure case, since it records no important objects. It is because the proposed method focuses on the global audio and visual features, while neglects the local object features. It inspires us to design an object-aware feature encoding scheme in the future work.

\section{Conclusions and Future Work}\label{section5}

In this paper, a hierarchical multimodal transformer is proposed for the video summarization task. It employs the transformer to capture the global dependency among frame sequence, which is more superior than the temporal dependency captured by RNN. According to the two-layered video structure, a hierarchical structure is developed by stacking two transformers. They process the frames-level dependency and shot-level dependency hierarchically. Furthermore, a multimodal fusion mechanism is designed to integrate the visual and audio information together to summarize the video. The results have demonstrated the superiority of the proposed method than existing traditional, RNN-based and attention-based methods, both in the performance and efficiency.

Audio and visual information are both important for the video summarization task. However, most videos suffer from the asynchronism problem of audio and visual information, which is not considered in this paper. Besides, object-aware features are essential to maintain important objects in the summarization process, which is neglected in this paper. In the future work, we plan to address the above problems by developing effective audiovisual registration strategies and object-aware feature encoding schemes, so as to promote the quality of video summary.

\section{Acknowledgement}\label{section6}
This work was supported in part by the National Key Research and Development Program of China under Grant 2018AAA0102200, in part by the Natural Science Basic Research Program of Shaanxi under Grant 2021JQ-204, in part by the China Postdoctoral Science Foundation under Grant 2020TQ0236.

\bibliography{neuro}

\end{document}